\documentclass[11pt]{article}

\usepackage[preprint]{acl}
\usepackage{times}
\usepackage{latexsym}

\usepackage[T1]{fontenc}

\usepackage[utf8]{inputenc}

\usepackage{microtype}

\usepackage{inconsolata}

\usepackage{graphicx}

\usepackage{url}            
\usepackage{booktabs}       
\usepackage{amsfonts}       
\usepackage{nicefrac}       
\usepackage{xcolor}         
\usepackage{algorithm}
\usepackage{algpseudocode}
\usepackage{subcaption}
\usepackage{array}
\newcolumntype{L}[1]{>{\raggedright\arraybackslash}p{#1}}
\usepackage{listings}
\lstdefinestyle{prompt}{
  basicstyle=\ttfamily\footnotesize,
  breaklines=true,
  breakatwhitespace=false,
  columns=fullflexible,
  keepspaces=true,
  frame=lines,
  xleftmargin=1em,
  xrightmargin=1em,
  aboveskip=0.6em,
  belowskip=0.9em,
  literate={×}{{$\times$}}1 {’}{{'}}1
}

\usepackage{amsmath,amsfonts,bm}

\def\eqref#1{equation~\ref{#1}}

\def\1{\bm{1}}

\def\vb{{\bm{b}}}

\def\vq{{\bm{q}}}
\def\vr{{\bm{r}}}

\def\vx{{\bm{x}}}

\DeclareMathAlphabet{\mathsfit}{\encodingdefault}{\sfdefault}{m}{sl}
\SetMathAlphabet{\mathsfit}{bold}{\encodingdefault}{\sfdefault}{bx}{n}

\def\gD{{\mathcal{D}}}

\def\gS{{\mathcal{S}}}

\def\sR{{\mathbb{R}}}

\graphicspath{ {./figures/} }
\usepackage{bbding}
\usepackage{wrapfig,lipsum}
\usepackage{amsmath}
\usepackage{amssymb}
\usepackage{mathtools}
\usepackage{amsthm}
\usepackage{multicol}

\theoremstyle{plain}
\newtheorem{theorem}{Theorem}[section]

\theoremstyle{definition}

\newtheorem{assumption}[theorem]{Assumption}
\theoremstyle{remark}

\usepackage[textsize=tiny]{todonotes}

\title{Revisiting What Makes a Good Context Subset: A Sparse Latent Basis Perspective}

\author{Cen-Jhih Li\\
  Kahlert School of Computing\\
  University of Utah\\
  Salt Lake City, UT 84112\\USA\\
  \texttt{cenjhih.li@gmail.com} \\\And
  Siddhant Chaudhary \\
  Kahlert School of Computing\\
  University of Utah\\
  Salt Lake City, UT 84112\\USA\\
  \texttt{urssidd@gmail.com}\And
  Aditya Bhaskara \\
  Kahlert School of Computing\\
  University of Utah\\
  Salt Lake City, UT 84112\\USA\\
  \texttt{bhaskaraaditya@gmail.com} \\}

\begin{document}
\maketitle
\begin{abstract}
Selecting context subsets is increasingly important in modern LLM inference, yet what makes a subset effective remains unclear. We study this question in the controlled setting of few-shot demonstration selection. We examine a hypothesis that demonstrations should collectively provide task-relevant information needed by a target query. Assuming that the query representation provides a useful proxy for these requirements, we formulate selection as sparse reconstruction and instantiate it with Residual Explanation Search (RES), a Matching-Pursuit-style diagnostic procedure. Across seven benchmarks and eight open-weight LLMs, no retrieval objective is universally best. RES outperforms MMR in 34 of 56 model--dataset pairs, with its clearest advantage on sentiment tasks, suggesting that query-conditioned collective reconstruction may capture useful complementarity that pairwise dissimilarity misses. Relevance-based retrieval remains strong on template-heavy reasoning benchmarks, while all retrieval methods underperform zero-shot prediction on average on StrategyQA, where a solution-reconstruction analysis is consistent with the query not exposing the required external knowledge. Target-model hidden states further show that retrieval objectives interact differently with representation spaces: middle-layer representations generally favor similarity retrieval, whereas last-layer representations are often stronger for residual- and diversity-based selection. These findings support a conditional, representation-dependent account of context subset quality rather than a universally superior selector.
\end{abstract}

\section{Introduction}\label{sec:intro}
Modern large language model (LLM) systems increasingly operate over large and complex contexts, making the selection of a small, task-relevant subset an important inference-time problem~\citep{anthropic2024claude3,singh2025openai,xu2024retrieval,zou2026latent, zhao2026retrieval}. We study this broader problem through demonstration selection for few-shot in-context learning (ICL;~\citealp{brown2020language,luo2024incontext,purohit2025sample, zhang2025linear,qi2026autoregressive}). This setting provides a controlled testbed in which candidate context items have a common structure and their effects on model predictions can be measured directly.

Even in this controlled setting, a fundamental question remains: \emph{what makes a good demonstration subset?} Prior work shows that ICL performance depends on demonstration content and ordering, while recent methods optimize retrieval policies, diversity, coverage, or prompt order~\citep{cobbina-zhou-2025-show,li2025order,chung2026manyshot,wu2024prompt,purohit2025sample,qi2026autoregressive,xin-etal-2026-ucs}. Yet it remains unclear what subset-level property these objectives approximate and what the selected demonstrations should collectively provide.

We consider one possible account: \emph{an effective demonstration subset should collectively provide the semantic and reasoning capabilities required to solve the target query}~\citep{chen2024skills}. This account is intuitive, but it is difficult to evaluate directly because those capabilities are not explicitly observable. We therefore operationalize it through a representation-space proxy. Under the assumption that the query representation contains useful information about the target instance's requirements, a good subset should collectively explain, or reconstruct, that representation. This leads to a sparse latent-basis hypothesis for demonstration selection, in which selected exemplars are viewed as a small set of basis elements whose linear combination approximates the target query representation.

To instantiate this hypothesis, we use a Matching-Pursuit-style~\citep{mallat1993matching} sparse approximation procedure for query-conditioned demonstration selection, which we call Residual Explanation Search (RES). RES iteratively selects an exemplar aligned with the remaining unexplained component of the query representation. Rather than presenting RES as an algorithmic contribution, we use it as a diagnostic procedure to test whether residual explanation is associated with improved ICL performance. This allows us to examine when observed ICL performance patterns are consistent with the hypothesis, when its underlying assumption appears less reliable, and how these patterns change across retrieval representations.

Our contributions are twofold. First, we formulate demonstration selection through the sparse latent-basis hypothesis and instantiate RES as a diagnostic of query-conditioned complementarity. Second, experiments across seven benchmarks and eight LLMs characterize when residual explanation aligns with ICL gains and reveal that retrieval objectives interact differently with representation spaces.

\section{Related Work}
\label{sec:relate}
\paragraph{Demonstration selection and subset properties.}
Demonstration selection has been widely studied for few-shot in-context learning. Methods range from lexical and semantic retrieval~\citep{robertson2009probabilistic,liu2022makes,margatina2023active} to approaches that optimize subset scores, selection policies, exemplar ordering, gradient- or influence-based signals, and reinforcement-learning objectives~\citep{purohit2024explora,luo2024incontext,wu2024prompt,purohit2025sample,zhang2025linear,wang2025demonstration,qi2026autoregressive}. Although these methods improve context construction, we focus on a complementary question: what subset-level property should an effective demonstration set satisfy?

A related line of work characterizes subset quality through complementarity and coverage. \citet{ye2023complementary} use MMR to trade off query relevance and pairwise dissimilarity, while \citet{pmlr-v202-ye23c} model query--demonstration interactions with a determinantal point process. Coverage-based approaches more directly model what a subset collectively contains. \citet{gupta-etal-2023-coverage} introduce set-level metrics for covering salient aspects of the target input. More recently, TopicK~\citep{kweon-etal-2025-topic} retrieves demonstrations to cover predicted topic requirements and model knowledge gaps, whereas UCS~\citep{xin-etal-2026-ucs} estimates unseen latent-cluster coverage. These approaches operationalize complementarity through pairwise dissimilarity, learned interactions, or coverage over salient content, topics, or clusters. We instead examine whether complementarity can be expressed as query-conditioned sparse reconstruction in a dense representation space.

\paragraph{Representation spaces for retrieval.}
Retrieval objectives depend critically on the representation space used to compare queries and exemplars. Sentence encoders such as Sentence-BERT~\citep{reimers-2019-sentence-bert} are commonly used as general-purpose semantic representations, while recent studies suggest that different LLM layers encode different kinds of information: intermediate layers often provide stronger semantic or transferable representations, whereas later layers are more closely tied to next-token prediction~\citep{skean2025layer}. Work on function vectors and linear representations further suggests that task-relevant information can sometimes be manipulated through directions or linear operations in activation space~\citep{ICLR2024_4ae163cb,park2024linear}. Motivated by this, we study how retrieval objectives behave under Sentence Transformer embeddings, middle-layer hidden states, and last-layer hidden states.

\section{Method}\label{sec:method}

\subsection{Motivation}
Recent work increasingly recognizes that demonstration quality depends not only on relevance, but also on diversity and coverage~\citep{chen2024skills,wang2025demonstration,xin-etal-2026-ucs}. While existing methods optimize these properties through different retrieval objectives, the underlying principle that makes a context subset effective remains unclear. To better understand why these retrieval objectives are effective, we consider one possible account~\citep{chen2024skills}:

\noindent
\emph{
An effective demonstration subset collectively provides the semantic and reasoning capabilities required to solve the target query.
}

Under this view, diverse or complementary demonstrations are useful because they contribute different pieces of task-relevant information, rather than repeatedly providing similar evidence. This perspective is not directly testable. Our goal is therefore to make the perspective operational in representation space and ask whether the resulting retrieval behavior is consistent with observed ICL performance.

\subsection{Problem Formulation}
The above perspective is conceptual, so we operationalize it through a retrieval formulation, allowing empirical evaluation of whether subsets consistent with this perspective lead to improved in-context learning performance. Specifically, we study context selection under the $k$-shot in-context learning (ICL) setting. Given the representation of a target query $\vq$ and a candidate pool of exemplar representations $\gD=\{\vx_1,\ldots,\vx_n\}$, our goal is to select a subset $\gS\subseteq\gD$ with $|\gS|=k$ whose demonstrations collectively provide what the target query requires.

\begin{assumption}[Query Representation]\label{assumption} The query representation provides a useful proxy for the semantic and reasoning requirements of the target task. \end{assumption}

Assumption~\ref{assumption} reflects the practical retrieval setting, where the target query is the only information available before selecting demonstrations. It is not expected to hold universally: when the query does not reveal the knowledge or reasoning process required for the task, retrieval based on the query representation may fail. We use this assumption as the starting point for the following formulation.

To formalize the problem, we start from modeling the query representation as a linear combination of latent components:
\begin{align}
    \vq=\sum_{j=1}^m\beta_j\vb_j,\label{eq:query}
\end{align}
where $\vb_j$ denotes the $j$-th latent task-relevant component and $\beta_j$ represents its contribution to the query. Conceptually, each latent component corresponds to one requirement of the target query.

Similarly, each candidate exemplar embedding $\vx_i\in\gD$ is represented as
\begin{align}
    \vx_i=\sum_{j=1}^m\gamma_{ij}\vb_j,\label{eq:exemplar}
\end{align}
where $\gamma_{ij}$ measures the contribution of the $j$-th semantic component to exemplar $\vx_i$. Although these latent task-relevant components are not directly observable, they provide a conceptual framework for understanding why diverse exemplars may benefit ICL.

The above latent decomposition naturally leads to the following formulation of exemplar selection. Here, to model the collective contribution of multiple demonstrations, we further assume that their contextual influences are approximately additive in the latent space~\citep{ICLR2024_4ae163cb,park2024linear,ilharco2023editing}. Under this assumption, the collective contribution of a selected subset can be approximated by a linear combination of its exemplar representations, leading to the following Latent-Basis Hypothesis.

\noindent\textbf{Latent Basis Hypothesis.}
\emph{
Given a target query representation $\vq$ and candidate exemplar representations $\gD$, represented according to Eqs.~\ref{eq:query} and~\ref{eq:exemplar}, the latent basis hypothesis models the target query representation as being approximated by the linear combination of a selected subset:
\[
\sum_{\vx_i\in\mathcal S}\alpha_i\vx_i
\approx\vq.
\]
}

This definition naturally leads to the following objective function:
\begin{align}
\min_{\gS\subseteq\gD,|\gS|=k,\alpha_i\in\sR}\lVert\vq- \sum_{\vx_i\in \gS}\alpha_i\vx_i\rVert. \label{eq:obj}
\end{align}
The resulting optimization problem is NP-hard~\cite{natarajan1995sparse}, so we use a greedy residual procedure to instantiate the hypothesis in our experiments.

\subsection{Residual Explanation Search (RES)}
To obtain a tractable, query-aligned heuristic for Eq.~\ref{eq:obj}, we instantiate a Matching-Pursuit-style~\citep{mallat1993matching} sparse approximation procedure for query-conditioned demonstration selection. We call the resulting procedure Residual Explanation Search (RES). We use it as a concrete probe of the latent-basis hypothesis.

At a high level, RES treats the unexplained portion of the query representation as a residual and retrieves against this residual iteratively. In this study, \emph{explanation} refers to geometric reconstruction of the query representation, not to a causal or human-interpretable explanation. Before retrieval, all $\vx_i$ and $\vq$ are L2-normalized, and the residual is initialized as $\vr=\vq$. At each iteration, RES selects the candidate with the largest signed inner product $\langle\vr,\vx\rangle$. Since candidate embeddings remain unit-normalized, this ranking is equivalent to cosine similarity with the current residual. Unlike classical Matching Pursuit, which ranks atoms by absolute correlation, RES uses signed alignment so that a strongly oppositely aligned exemplar is not treated as a useful match merely because a negative coefficient could reverse its direction. After selection, RES sets $\alpha=\langle\vr,\vx^\star\rangle$ and subtracts the corresponding projection from the residual. Here, $\alpha$ is an internal reconstruction coefficient rather than a weight applied to the demonstration in the ICL prompt. Repeating this process encourages later selections to account for portions of the query representation not explained by earlier examples, thereby modeling complementarity through residual explanation rather than pairwise exemplar dissimilarity. Algorithm~\ref{alg:res} summarizes the procedure.

\begin{algorithm}[t]
\caption{Residual Explanation Search (RES)}
\label{alg:res}
\begin{algorithmic}[1]
\Require
Query embedding $\vq$,
candidate set $\gD$,
number of exemplars $k$
\Ensure
Selected exemplar subset $\gS$

\State $\vr \gets \vq$ \Comment{Initialize residual}
\State $\gS \gets \emptyset$

\For{$i=1$ to $k$}

    \State
    $\vx^\star
=
\arg\max_{\vx\in\gD\setminus\gS}
\langle\vr,\vx\rangle$

    \Comment{Largest signed residual alignment}

    \State
    $\gS \gets \gS \cup \{\vx^\star\}$

    \State
    $\alpha \gets \langle \vr, \vx^\star\rangle$

    \State
    $\vr \gets \vr-\alpha \vx^\star$

    \Comment{Subtract explained component}

\EndFor

\State \Return $\gS$

\end{algorithmic}
\end{algorithm}
Under exhaustive search, RES requires $O(nkd)$ online computation for $n$ candidate exemplars, embedding dimension $d$, and $k$ selected demonstrations. Details are provided in Appendix~\ref{apdx:complexity}. We next examine how representation choice affects the information available for retrieval.

\subsection{Representation for Exemplar Retrieval}\label{method:representation}

Our formulation relies on Assumption~\ref{assumption}, which may not hold equally well across different representation spaces. Consequently, the choice of representation becomes an important question, as different representations may expose these requirements to different extents.

To understand how different representation spaces interact with retrieval objectives, we compare embeddings from \texttt{all-MiniLM-L6-v2}, used as a general-purpose Sentence Transformer baseline~\citep{reimers-2019-sentence-bert}, with hidden representations extracted from the target LLM. Sentence Transformer embeddings are commonly used as general-purpose semantic representations. We further consider two internal representations of the target LLM, each providing a different view of the information available to the retrieval objective.

Prior work suggests that middle-layer representations better capture semantic information and transfer more effectively across downstream tasks~\citep{skean2025layer}, whereas late-layer representations become increasingly specialized for next-token prediction. These two representation spaces therefore naturally correspond to different assumptions about the information required for effective retrieval. Middle-layer representations allow us to test retrieval in \emph{a more semantic representation space}, whereas last-layer representations test retrieval in \emph{a space closer to the model's prediction process}.

For hidden-state retrieval, we convert each input into a single vector by mean pooling token-level activations. Middle-layer representations are obtained by averaging token hidden states from the second quartile of the network depth (e.g., layers 9--16 in a 32-layer model), while last-layer representations are obtained by averaging token hidden states from the final layer. We use the same pooling procedure for queries and candidate exemplars, and compare Sentence Transformer embeddings, middle-layer hidden states, and last-layer hidden states under identical retrieval objectives throughout our experiments.

\begin{table*}[t]
\centering
\small
\begin{tabular}{lccccccc}
\toprule
Method & AQUA & MathQA & QNLI & SST-2 & SST-5 & StrategyQA & SUBJ \\
\midrule
random & 8/8 (100.0\%) & 8/8 (100.0\%) & 5/8 (62.5\%) & 7/8 (87.5\%) & 8/8 (100.0\%) & 5/8 (62.5\%) & 8/8 (100.0\%) \\
bm25 & 1/8 (12.5\%) & 0/8 (0.0\%) & 2/8 (25.0\%) & 5/8 (62.5\%) & 5/8 (62.5\%) & 4/8 (50.0\%) & 2/8 (25.0\%) \\
top-$k$ & 1/8 (12.5\%) & 0/8 (0.0\%) & 3/8 (37.5\%) & 6/8 (75.0\%) & 4/8 (50.0\%) & 1/8 (12.5\%) & 1/8 (12.5\%) \\
mmr & 3/8 (37.5\%) & 5/8 (62.5\%) & 4/8 (50.0\%) & 6/8 (75.0\%) & 7/8 (87.5\%) & 3/8 (37.5\%) & 6/8 (75.0\%) \\
\bottomrule
\end{tabular}
\caption{Number (percentage) of the eight LLMs for which RES achieves higher accuracy than each baseline. Random results are averaged over three seeds.}
\label{tab:win_rate}
\end{table*}

\begin{table*}[t]
\centering
\small
\begin{tabular}{l|ccccccc}
\toprule
task & AQUA & MathQA & QNLI & SST-2 & SST-5 & SUBJ & StrategyQA \\
method &  &  &  &  &  &  &  \\
\midrule
bm25 & $6.99_{\pm 3.41}$ & $11.66_{\pm 7.24}$ & $6.50_{\pm 6.77}$ & $5.37_{\pm 6.19}$ & $2.18_{\pm 5.31}$ & $20.99_{\pm 9.42}$ & $-1.84_{\pm 4.15}$ \\
mmr & $5.61_{\pm 3.40}$ & $4.63_{\pm 4.03}$ & $5.49_{\pm 5.99}$ & $5.63_{\pm 6.06}$ & $3.07_{\pm 5.78}$ & $16.46_{\pm 8.78}$ & $-1.78_{\pm 3.60}$ \\
random & $3.07_{\pm 4.18}$ & $2.71_{\pm 4.06}$ & $5.54_{\pm 6.18}$ & $3.40_{\pm 6.43}$ & $-1.14_{\pm 5.74}$ & $7.46_{\pm 7.87}$ & $-1.79_{\pm 3.90}$ \\
res & $4.77_{\pm 3.93}$ & $4.76_{\pm 3.87}$ & $5.59_{\pm 6.31}$ & $6.10_{\pm 5.27}$ & $3.19_{\pm 5.51}$ & $17.26_{\pm 9.43}$ & $-1.86_{\pm 4.07}$ \\
top-$k$ & $7.58_{\pm 5.45}$ & $11.18_{\pm 7.38}$ & $6.34_{\pm 6.55}$ & $5.98_{\pm 6.49}$ & $3.69_{\pm 5.26}$ & $23.61_{\pm 10.08}$ & $-1.24_{\pm 3.94}$ \\
\bottomrule
\end{tabular}
\caption{Mean accuracy gain over zero-shot across eight LLMs (mean $\pm$ standard deviations). The random baseline is averaged over three seeds for each LLM (24 runs).}
\label{tab:mean_gain}
\end{table*}

\section{Experiment Settings}\label{sec:exp}

We evaluate seven benchmarks covering sentiment and subjectivity classification (SST-2 and SST-5~\citealp{socher2013recursive}; SUBJ~\citealp{pang2004sentimental}), natural language inference (QNLI~\citealp{wang2018glue}), and knowledge-intensive and mathematical reasoning (StrategyQA~\citealp{geva2021did}, AQUA~\citealp{ling2017program}, and  MathQA~\citealp{amini2019mathqa}). All experiments use 5-shot ICL with eight open-weight LLMs: Gemma-2B-IT~\citep{gemmateam2024gemma}, Llama-3.2-3B-Instruct and Llama-3.1-8B~\citep{grattafiori2024llama}, Qwen2.5-3B-Instruct and Qwen2.5-7B~\citep{qwen2024qwen25}, Phi-3-Mini-128K-Instruct and Phi-3.5-Mini-Instruct~\citep{abdin2024phi3}, and DeepSeek-LLM-7B-Base~\citep{deepseekai2024deepseekllm}. We use direct label-prediction prompts for classification and rationale-augmented question-answer prompts for reasoning. Templates are provided in Appendix~\ref{apdx:prompt}.

We compare RES with BM25~\citep{robertson2009probabilistic}, Top-$k$ semantic retrieval~\citep{liu2022makes}, MMR~\citep{ye2023complementary}, and random selection, representing lexical relevance, semantic relevance, relevance with pairwise dissimilarity, and selection without optimization. We focus on these representative objectives to isolate whether residual explanation captures a property distinct from relevance and pairwise diversity, without introducing additional reranking, coverage, learning, or ordering components~\citep{wang2025demonstration,purohit2025sample,pham2025rapid,xin-etal-2026-ucs}. For random selection, we report mean performance over three seeds. Other implementation details can be found in Appendix~\ref{apdx:implementation}.

\section{Results and Discussion}\label{sec:result}
We first present the overall behavior of RES across datasets and LLMs, focusing on when residual explanation improves over standard relevance- and diversity-based retrieval objectives. We then analyze representative success and failure cases to clarify the boundary conditions of the latent-basis perspective. Finally, we study how retrieval representations interact with different retrieval objectives.

\subsection{When Does Residual Explanation Help?}
Tables~\ref{tab:win_rate} and~\ref{tab:mean_gain} show that there is no universally best retrieval objective across datasets. Relevance-based methods such as Top-$k$ and BM25 remain strong on several benchmarks, especially MathQA, AQUA, and SUBJ. RES outperforms random selection in 49 of 56 model-dataset pairs and MMR in 34 of 56 pairs. The contrast with MMR is clearest on SST-2 and SST-5, where RES outperforms MMR on six and seven of the eight evaluated LLMs, respectively, while remaining competitive with relevance-based retrieval. These results suggest that query-conditioned reconstruction captures a subset property distinct from pairwise dissimilarity and may provide a more useful proxy for complementarity in these settings.

However, residual explanation is not uniformly better than relevance-based retrieval. On MathQA and AQUA, Top-$k$ and BM25 often achieve stronger average gains. One plausible explanation is that many instances in these benchmarks follow reusable templates or near-duplicate solution patterns, so retrieving the most similar demonstrations may already provide the relevant procedure. StrategyQA shows a different empirical pattern, where all retrieval methods underperform zero-shot prediction on average. We therefore use an additional reconstruction analysis to examine whether Assumption~\ref{assumption} is less reliable in these settings.

To further examine this interpretation, we conduct a solution reconstruction analysis on the three reasoning benchmarks. After each retrieval method selects demonstrations from query-only representations, we augment each target question with its gold answer and rationale or supporting facts and measure how well the selected demonstrations reconstruct this fuller solution representation. We use the resulting score to diagnose the reliability of Assumption~\ref{assumption}, not to compare retrieval objectives. Calculation details are provided in  Appendix~\ref{apdx:solution_recon}.

\begin{table}[t]
\centering
\small
\setlength{\tabcolsep}{4pt}
\begin{tabular}{lcccc}
\toprule
Dataset & RES & Top-$k$ & MMR & Random \\
\midrule
MathQA     & $.904_{\pm .102}$ & $.914_{\pm .111}$ & $.906_{\pm .101}$ & $.276_{\pm .109}$ \\
AQUA       & $.744_{\pm .125}$ & $.710_{\pm .160}$ & $.748_{\pm .126}$ & $.259_{\pm .104}$ \\
StrategyQA & $.364_{\pm .107}$ & $.353_{\pm .112}$ & $.347_{\pm .110}$ & $.048_{\pm .035}$ \\
\bottomrule
\end{tabular}
\caption{Solution reconstruction scores on reasoning benchmarks for demonstrations selected by RES, Top-$k$, MMR, and random selection, using Sentence Transformer embeddings. Values are means $\pm$ standard deviations across target instances. Random selection pools three seeds.}
\label{tab:solution_recon}
\end{table}

The reconstruction analysis provides a dataset-level view of when Assumption~\ref{assumption} appears reliable. Table~\ref{tab:solution_recon} shows a substantially lower reconstruction score on StrategyQA than on MathQA and AQUA. This ordering holds for all three retrieval methods, whose scores differ by at most 0.04 within each dataset. All three also score far above random selection, and their gain over random selection is smallest on StrategyQA (at most 0.32, compared with at least 0.45 on MathQA and AQUA), suggesting that the gap reflects the datasets rather than the RES objective. In particular, Assumption~\ref{assumption} appears less reliable on StrategyQA, where the query often does not expose enough information for retrieval to identify demonstrations containing the required external facts. By comparison, the higher scores on MathQA and AQUA suggest that their queries expose more recoverable solution structure. This does not imply that residual explanation should outperform relevance-based retrieval on these datasets. When reusable templates are common, semantic relevance may already retrieve demonstrations that provide the main solution procedure. We next examine representative cases to further illustrate these dataset-level patterns.

\subsection{Case Discussion}\label{sec:cases}

\begin{table*}[t]
\small
\centering
\setlength{\tabcolsep}{4pt}
\renewcommand{\arraystretch}{1.12}
\begin{tabular}{L{0.10\textwidth} L{0.28\textwidth} L{0.14\textwidth} L{0.41\textwidth}}
\toprule
Dataset & Target query & Comparison & Retrieved-context pattern \\
\midrule
StrategyQA &
Was ship that recovered Apollo 13 named after a World War II battle? &
Top-$k$ vs. random &
Top-$k$ retrieves space-related demonstrations involving \textbf{Apollo missions}, the \textbf{moon}, and \textbf{lunar landing}. Random retrieves unrelated facts. Neither provides the missing facts about the Apollo 13 recovery ship or its name origin. \\
\midrule
SST-2 &
A fairly by-the-books blend of action and romance with sprinklings of intentional and unintentional comedy. &
RES vs.\ \mbox{Top-$k$/MMR} &
Top-$k$ focuses on \textbf{romantic comedy} examples, and MMR adds mostly topical or stylistic variation. RES retrieves more varied sentiment cues, such as \textbf{odd}, \textbf{haphazard}, \textbf{serious suspense}, and \textbf{uniformly good}. \\
\midrule
MathQA &
A man bought 20 shares of Rs. 50 at 5 discount, the rate of dividend being 13. The rate of interest obtained is: &
RES vs. MMR &
Both methods retrieve a near-template match involving \textbf{shares}, \textbf{discount}, \textbf{dividend}, and \textbf{interest rate}. The remaining demonstrations differ but are less clearly decisive. \\
\midrule
AQUA &
Let A and B be independent events with $P(A)=0.2$ and $P(B)=0.8$. Find $P(A/B)$. &
RES vs. MMR &
MMR retrieves several \textbf{probability}-related demonstrations, while RES retrieves broader algebra, ratio, business, and logarithm examples. \\
\bottomrule
\end{tabular}
\caption{Representative retrieval cases. We show abridged retrieval patterns rather than full prompts. Complete prompts are provided in Appendix~\ref{apdx:case_prompts}.}
\label{tab:case_examples}
\end{table*}

The retrieval setting considered in this work depends on Assumption~\ref{assumption}, which treats the query representation as a proxy for the target instance’s requirements. When the query does not expose them, retrieval may select superficially related demonstrations that do not provide the knowledge or reasoning needed for the target instance. Table~\ref{tab:case_examples} summarizes representative cases used in the following discussion. Full prompts are provided in Appendix~\ref{apdx:case_prompts}.

We first examine StrategyQA, where all retrieval methods underperform zero-shot prediction on average. Table~\ref{tab:case_examples} presents a representative example comparing Top-$k$ and random selection, with full prompts provided in Appendix~\ref{apdx:case_prompts}. In this example, Top-$k$ retrieves demonstrations involving Apollo missions, the moon, and lunar landing, which are semantically related to the target query. Nevertheless, the target question requires a more specific chain of knowledge about the ship that recovered Apollo 13 and whether its name refers to a World War II battle. The retrieved demonstrations share the surface topic of space exploration, but do not provide this missing knowledge. This illustrates why semantic similarity can fail to improve ICL when the query representation does not expose the required reasoning path.

In contrast, SST-2 illustrates a setting where the query more directly exposes label-relevant information. The target review contains topical cues such as action, romance, and comedy, but its sentiment depends on evaluative expressions such as ``fairly by-the-books.'' In the representative example in Table~\ref{tab:case_examples}, Top-$k$ retrieves demonstrations that are strongly tied to the romance-comedy topic, while MMR introduces pairwise diversity but still largely operates within related topical or stylistic neighborhoods. RES instead retrieves demonstrations containing a broader set of sentiment-bearing expressions, including both negative and positive evaluative cues. This example is consistent with the view that residual explanation can help when the query representation contains observable cues relevant to the target prediction.

Finally, MathQA and AQUA represent intermediate and less diagnostic cases. Many questions in these benchmarks share reusable mathematical templates, formulas, or near-duplicate solution patterns. In the MathQA example in Table~\ref{tab:case_examples}, both RES and MMR retrieve a highly similar stock-dividend-interest demonstration, which likely provides the most relevant solution template. The remaining selected demonstrations differ, but their effect on the final prediction is harder to attribute. Conversely, in the AQUA example, MMR retrieves more probability-related demonstrations and succeeds, while RES retrieves a broader mixture of algebra, ratio, business, and logarithm examples and fails. These examples suggest that, for template-heavy reasoning datasets, relevance-based retrieval can already capture much of the useful information, reducing the additional benefit of explicitly modeling complementary latent components.

\subsection{Representation Study}

\begin{figure*}[t]
    \centering

    \begin{subfigure}{0.95\textwidth}
        \centering
        \includegraphics[width=\linewidth]{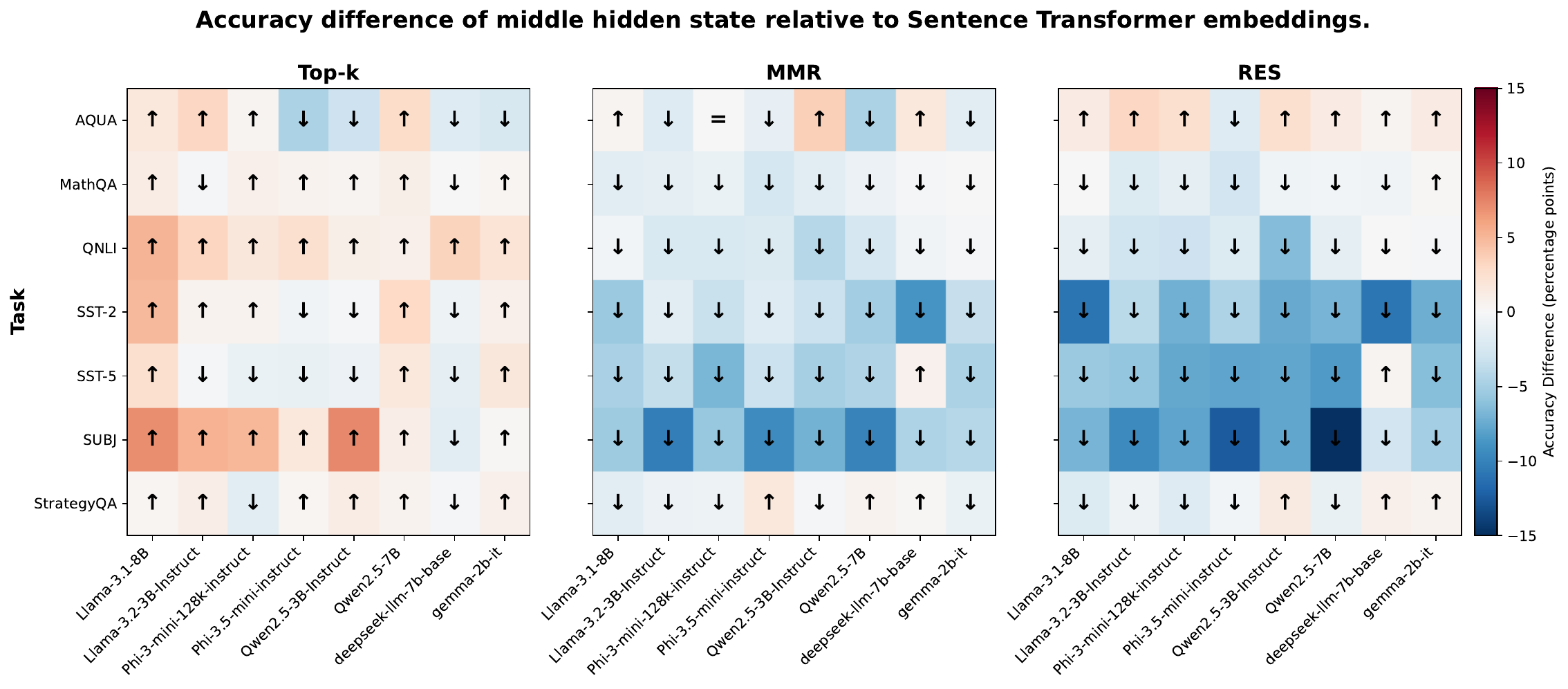}
        \caption{Middle-layer hidden state representations.}
    \end{subfigure}

    \vspace{0.8em}

    \begin{subfigure}{0.95\textwidth}
        \centering
        \includegraphics[width=\linewidth]{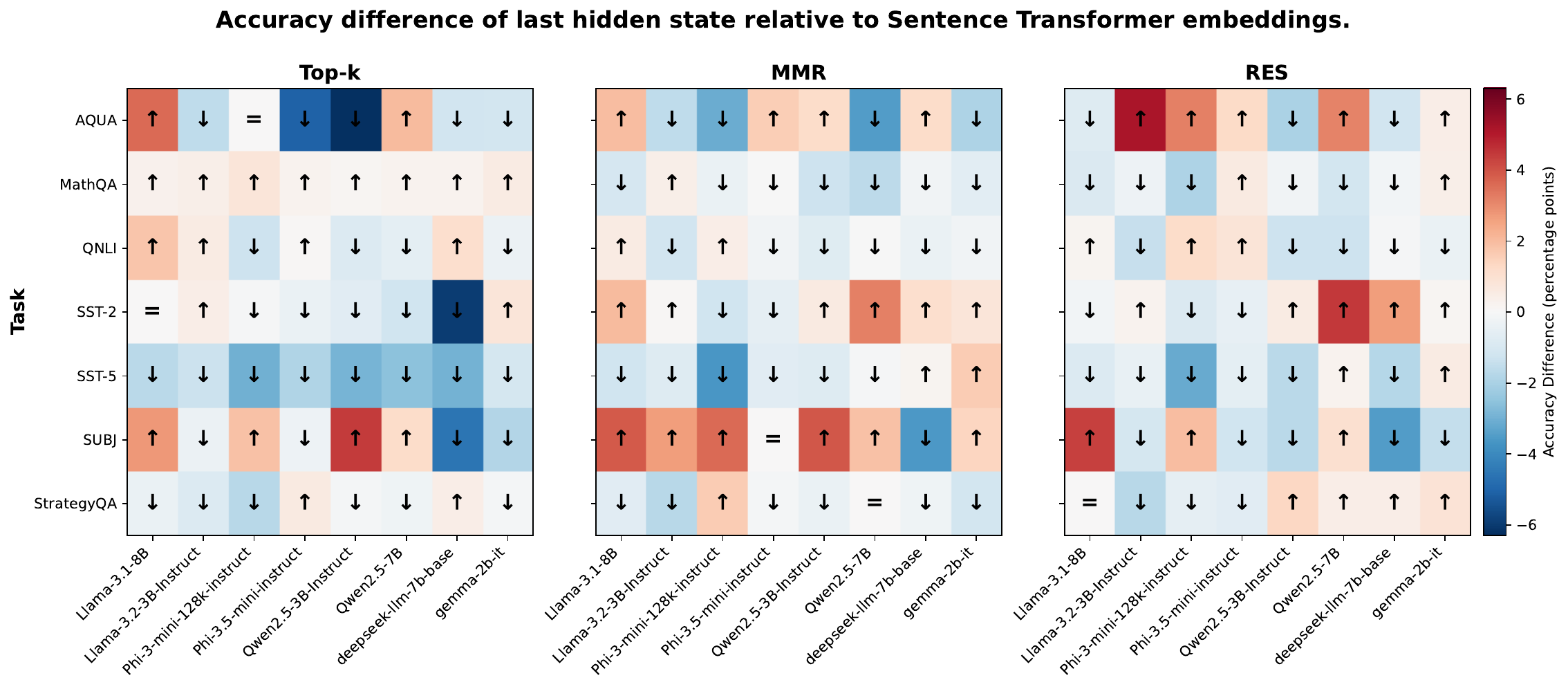}
        \caption{Last-layer hidden state representations.}
    \end{subfigure}

    \caption{Accuracy difference (percentage points) relative to Sentence Transformer embeddings when retrieval instead uses hidden representations extracted from the target LLM. Positive values indicate improvements. Arrows show the sign of the difference, and $=$ denotes no change. The two rows use different color scales.}\label{fig:representation_comparison}
\end{figure*}
Having examined how retrieval objectives behave across datasets, we now investigate how their behavior changes across representation spaces. Figure~\ref{fig:representation_comparison} reports the accuracy difference relative to Sentence Transformer embeddings when retrieval instead uses hidden representations from the target LLM. As discussed in Section~\ref{method:representation}, we use middle-layer and last-layer hidden states as approximate views of semantic and prediction-oriented information, respectively.

These retrieval objectives lead to different expectations about representation choice. Top-$k$ constructs demonstration subsets according to semantic similarity to the query, while MMR additionally promotes pairwise semantic diversity among the selected demonstrations. If semantic similarity is indeed the primary information used by these objectives, replacing Sentence Transformer embeddings with the target LLM's middle-layer hidden states would be expected to preserve, or potentially improve, retrieval performance, as prior work suggests that middle-layer representations tend to retain more general semantic information than other layers within the target LLM~\citep{skean2025layer}. RES, in contrast, does not treat semantic similarity itself as the retrieval objective. Under the latent-basis hypothesis, it relies on Assumption~\ref{assumption}. Consequently, there is no clear expectation that a more semantically oriented representation should necessarily improve RES. Instead, because last-layer representations are more closely aligned with the target model's prediction process, they may better preserve information related to its solution process, making them a potentially more suitable representation space for RES.

The results are broadly consistent with this expectation for Top-$k$, but not for MMR. Middle-layer representations maintain or improve Top-$k$ performance on most datasets. For MMR, however, middle-layer representations generally reduce performance, while last-layer representations maintain or improve performance on several datasets. This contrast suggests that MMR is particularly sensitive to representation choice and that pairwise semantic dissimilarity may not always correspond to useful complementarity. MMR's relatively stronger performance with last-layer representations is also consistent with the possibility that useful complementarity depends partly on prediction-related information.

RES exhibits a pattern closer to MMR than to Top-$k$. Last-layer representations generally yield stronger performance than middle-layer representations, although the latter are expected to retain more general semantic information. This pattern does not validate the latent-basis hypothesis, but it is consistent with the possibility that residual explanation may benefit from representations closer to the target model's prediction process. These observations suggest that the effect of representation choice depends on the retrieval objective, and that more semantically oriented representations are not uniformly preferable for context selection.

\section{Conclusion}\label{sec:conclusion}
We revisited demonstration selection through the hypothesis that an effective context subset should collectively provide the task-relevant information required by a target query. We operationalized this hypothesis as a sparse latent-basis formulation and instantiated it with RES as a diagnostic residual-selection procedure. Across seven benchmarks and eight open-weight LLMs, no retrieval objective is universally strongest. RES frequently improves over pairwise-diversity retrieval, particularly on sentiment tasks, whereas relevance-based methods remain effective when near-template demonstrations are available. StrategyQA further illustrates a limitation of query-conditioned selection when the query does not expose the required external knowledge. Our representation study additionally shows that retrieval objectives cannot be considered independently of their representation spaces. These results do not establish that LLMs combine demonstrations linearly or that residual explanation is universally optimal. Rather, they suggest that residual reconstruction provides a useful operational proxy for complementarity when the query representation contains recoverable task-relevant information.

\section*{Limitations}
Our formulation depends on the assumption that the query representation exposes the requirements of the target instance (Assumption~\ref{assumption}). This assumption may be unreliable for knowledge-intensive questions whose solution depends on facts or reasoning paths not evident from the query. Moreover, the linear latent-basis formulation is an operational approximation rather than a mechanistic claim about how LLMs combine demonstrations. Our experiments therefore establish consistency with this formulation in some settings, but do not identify explicit reasoning skills or demonstrate that contextual influence is inherently linear.

Our empirical scope is limited to seven English benchmarks, a 5-shot setting, and eight open-weight LLMs between approximately 2B and 8B parameters. We do not evaluate long-context or agent settings, frontier models, learned selectors, or other shot counts. The representation study uses mean pooling, a fixed middle-layer range, the final hidden layer, and one Sentence Transformer model. Alternative encoders and pooling choices may yield different behavior. The solution-reconstruction analysis additionally uses gold answers and rationales and can be affected by dataset redundancy, rationale style, candidate-pool coverage, and embedding geometry. Finally, the qualitative cases are illustrative rather than systematically sampled.

\section*{Ethical Considerations}

This work uses existing public benchmarks and pretrained open-weight language models. It does not collect personal data, involve human participants, or train new foundation models. Its primary foreseeable environmental impact is the computation required to run inference across multiple models and retrieval settings. In addition, retrieved demonstrations and model predictions may inherit biases or harmful content from the underlying datasets and pretrained models.



\bibliography{references}

\clearpage
\appendix
\section*{Appendix}
\addcontentsline{toc}{section}{Appendix}
\section{Complexity Comparison}\label{apdx:complexity}

\begin{table}[t]
\centering
\begin{tabular}{lccc}
\toprule
Method & Precompute & Run Time & Space \\
\midrule
Top-$k$ &
- &
$O(nd)$ &
$O(nd)$ \\

MMR &
- &
$O(nkd)$ &
$O(nd)$ \\

MMR$^\dagger$ &
$O(n^2d)$ &
$O(nk)$ &
$O(n^2)$ \\

RES &
- &
$O(nkd)$ &
$O(nd)$ \\
\bottomrule
\end{tabular}
\caption{Theoretical time and space complexity of representative retrieval strategies under exhaustive search. Here, $n$ is the number of candidate exemplars, $d$ is the embedding dimension, and $k$ is the number of selected demonstrations. Precompute denotes additional offline preprocessing beyond computing and storing candidate embeddings. MMR denotes an incremental implementation that updates candidate-to-selected-set scores after each selection, whereas MMR$^\dagger$ denotes an implementation with cached pairwise exemplar similarities.}
\label{tab:complexity}
\end{table}

Table~\ref{tab:complexity} summarizes the online retrieval cost and storage requirements of the representative retrieval strategies considered in this work. Top-$k$ retrieval requires a single pass over the candidate pool, giving $O(nd)$ online computation and $O(nd)$ storage for the candidate embeddings.

RES has linear memory usage and $O(nkd)$ online computation under exhaustive search. It performs $k$ residual-matching steps. At each step, it computes similarities between the current residual and all remaining candidates, then applies an $O(d)$ residual update. Since RES only uses query and candidate embeddings, its storage cost is $O(nd)$.

MMR has the same online order as RES when implemented incrementally without pairwise-similarity caching. After each selected exemplar is added to the subset, the similarity between that exemplar and each remaining candidate is computed once, and each candidate's current diversity penalty is updated. This gives $O(nkd)$ online computation and $O(nd)$ storage. Alternatively, if all pairwise exemplar similarities are precomputed and cached, MMR$^\dagger$ reduces online retrieval to $O(nk)$, at the expense of $O(n^2d)$ preprocessing and $O(n^2)$ memory.

\section{Prompt Description}\label{apdx:prompt}

For all tasks, demonstrations are formatted with gold labels and, when available, rationales or explanations. The target query follows the same input format but leaves the answer field blank for model prediction.

\begin{table}[ht]
\centering
\small
\begin{tabular}{ll}
\toprule
Dataset & Prompt Template \\
\midrule
SST-2 &
\texttt{Review: <review>}\\
&
\texttt{Sentiment: <positive|negative>}\\
\midrule
SST-5 &
\texttt{Review: <review>}\\
&
\texttt{Sentiment: <terrible|bad|okay|good|great>}\\
\midrule
SUBJ &
\texttt{Input: <text>}\\
&
\texttt{Type: <objective|subjective>}\\
\midrule
QNLI &
\texttt{<premise>}\\
&
\texttt{Can we know <hypothesis>? <Yes|No>}\\
\bottomrule
\end{tabular}
\caption{Prompt templates for classification tasks. The exemplars include the label after the second line, while target queries leave the label field blank for prediction.}
\end{table}

\begin{table}[ht]
\centering
\small
\setlength{\tabcolsep}{3pt}
\renewcommand{\arraystretch}{1.05}
\begin{tabular}{p{0.23\columnwidth} p{0.70\columnwidth}}
\toprule
Dataset & Prompt template \\
\midrule
StrategyQA &
\begin{tabular}[t]{@{}l@{}}
\textbf{Exemplar:} Question: <question>\\
Answer: <answer>\\
Reasoning: <facts>\\
\textbf{Query:} Question: <question>\\
Answer:
\end{tabular}
\\
\midrule
\begin{tabular}[t]{@{}l@{}}
MathQA\\
AQUA
\end{tabular} 
&
\begin{tabular}[t]{@{}l@{}}
\textbf{Exemplar:} Question: <question>\\
Options: <options>\\
Answer: <choice>\\
Explanation: <rationale>\\
\textbf{Query:} Question: <question>\\
Options: <options>\\
Answer:
\end{tabular}
\\
\bottomrule
\end{tabular}
\caption{Prompt templates for reasoning tasks.}
\label{tab:reasoning_prompts}
\end{table}

\section{Implementation Details}\label{apdx:implementation}
For each dataset, demonstrations are retrieved from the training split, and evaluation is conducted on the test split. We use the validation split when test labels are unavailable. We set the MMR trade-off coefficient to $\lambda=0.5$, following the value recommended in the original formulation. For BM25, Top-$k$, MMR, and RES, demonstrations are placed in retrieval order. Random demonstrations retain their sampled order. For the retrieval-based methods, the first demonstration is the highest-scoring candidate, while subsequent demonstrations follow each method's ranking or sequential objective. Our implementation is based on the OpenICL codebase~\citep{wu2023openicl}.

Retrieval embeds only the input text: the review or sentence for SST-2, SST-5, and SUBJ; the question and sentence for QNLI; the question for StrategyQA; and the question with its answer options for AQUA and MathQA. Gold labels and rationales appear only in the demonstrations shown to the LLM. Candidate pools are the full training splits, with 6,920, 8,544, 8,000, 104,743, 1,603, 97,467, and 29,837 examples for SST-2, SST-5, SUBJ, QNLI, StrategyQA, AQUA, and MathQA, respectively, evaluated on 1,821, 2,210, 2,000, 5,463, 687, 254, and 2,985 target instances. Following OpenICL’s PPLInferencer, the model scores each candidate label by conditional perplexity and predicts the label with the lowest perplexity.

\section{Solution Reconstruction Analysis}
\label{apdx:solution_recon}
We provide the formal definition of the solution reconstruction analysis introduced in the main text. RES, Top-$k$, and MMR select demonstrations using query-only representations, and random selection ignores the query. Gold solution information is used only in this post-retrieval analysis.

For each target instance, let $\tilde{\vq}$ denote the embedding of the target question augmented with its gold answer and rationale or supporting facts. Similarly, let $\tilde{\vx}_i$ denote the embedding of a selected demonstration containing its question, answer, and rationale or supporting facts. The tilde distinguishes these solution-augmented embeddings from the query-only representations used for retrieval.

Given a selected subset $\gS$, we define the normalized least-squares reconstruction score as
\[
\operatorname{Recon}(\gS)
=
1 -
\frac{
\min_{\alpha \in \mathbb{R}^{|\gS|}}
\left\|
\tilde{\vq}
-
\sum_{\vx_i \in \gS}
\alpha_i \tilde{\vx}_i
\right\|_2^2
}{
\left\|\tilde{\vq}\right\|_2^2
},
\] 
where $\alpha=(\alpha_i)_{\vx_i\in\gS}\in \mathbb{R}^{|\gS|}$ denotes the reconstruction coefficients. We solve the inner minimization using unconstrained ordinary least squares. All solution-augmented representations are encoded using \texttt{all-MiniLM-L6-v2} and then L2-normalized. Because the zero coefficient vector is feasible, the score lies in $[0,1]$ up to numerical precision. Higher values indicate a better linear reconstruction of the solution-augmented target representation.

We interpret this score only as a dataset-level diagnostic, not as a retrieval objective or a predictor of per-example correctness. A low score is consistent with Assumption~\ref{assumption} being less reliable, although it may also reflect limitations of the candidate pool, representation space, or linear reconstruction model.

\section{Full Prompts for Case Study}\label{apdx:case_prompts}

This appendix provides the complete retrieved demonstrations for the representative cases discussed in Section~\ref{sec:cases}. The actual prompt templates used in evaluation are given in Appendix~\ref{apdx:prompt}. For readability, each block below separates the gold label from the model output and writes the final target response as \texttt{Prediction: <prediction>} rather than the original prompt field \texttt{Answer:}. This formatting is only for presentation and is intended to distinguish model predictions from gold labels and demonstration answers.

\subsection{StrategyQA}
\paragraph{Gold label.} True.

\paragraph{Top-$k$ retrieved prompt.}\mbox{}\\[-0.8em]
\begin{lstlisting}[style=prompt]
Question: Were deaths from Apollo 13 mission eclipsed by other space missions?
Answer: True
Reasoning: Apollo 13 was a failed Moon landing in which the crew had faulty oxygen resources. All of the Apollo 13 crew returned safely to earth. The Challenger was a space shuttle that malfunctioned, killing all seven passengers. The Space Shuttle Columbia disaster had seven casualties.
Question: Could largest ship from Voyages of Christopher Columbus haul Statue of Liberty?
Answer: False
Reasoning: The largest ship Christopher Columbus used was the Santa Maria. The Santa Maria has a cargo capacity of 108 tons. The Statue of Liberty weighs 225 tons.
Question: Did the crew of Apollo 15 take pictures of Mount Sharp?
Answer: False
Reasoning: Mount Sharp is a mountain on Mars. The crew of Apollo 15 landed on the Moon, not Mars. No humans have ever landed on Mars.
Question: Is the Royal Air Force ensign on the moon?
Answer: False
Reasoning: The Royal Air Force ensign is the flag of the Royal Air Force The Royal Air Force is a branch of the British Armed Forces Britain has never landed on the moon
Question: Was Amazon involved in the lunar landing?
Answer: False
Reasoning: The lunar landing occurred in 1969. Amazon was founded in 1994.
Question: Was ship that recovered Apollo 13 named after a World War II battle?
Prediction: False
\end{lstlisting}

\paragraph{Random retrieved prompt.}\mbox{}\\[-0.8em]
\begin{lstlisting}[style=prompt]
Question: Are there enough people in the Balkans to match the population of Japan?
Answer: False
Reasoning: There are approximately 55 million people in the Balkans There are more than 125 million people in Japan
Question: Would Bandy be likely to become popular in Texas?
Answer: False
Reasoning: The American Bandy Association governs major Bandy play in the United States. There are no teams from Texas registered with the American Bandy Association. Sports involving ice and snow tend to be more popular in cold climates. Texas has an extremely hot climate.
Question: Do onions have a form that resembles the inside of a tree?
Answer: True
Reasoning: When bisected, an onion has rings that extend from the core to the outside. Trees are formed of a series of rings that extend from the inside to the outside
Question: Did Jack Dempsey fight the current WBC heavyweight champion?
Answer: False
Reasoning: Jack Dempsey died in 1983 The current WBC heavyweight champion is Tyson Fury Tyson Fury was born in 1988
Question: Would half muggle wizards fear Lord Voldemort?
Answer: True
Reasoning: Lord Voldemort created a slur to describe half muggle wizards, calling them mudbloods. Lord Voldemort sought to purge the wizarding world of half muggle wizards through death.
Question: Was ship that recovered Apollo 13 named after a World War II battle?
Prediction: False
\end{lstlisting}

\subsection{SST-2}
\paragraph{Gold label.} Positive.

\paragraph{RES retrieved prompt.}\mbox{}\\[-0.8em]

\begin{lstlisting}[style=prompt]
Review: an odd , haphazard , and inconsequential romantic comedy . Sentiment: negative
Review: although it lacks the detail of the book , the film does pack some serious suspense . Sentiment: positive
Review: the performances are uniformly good . Sentiment: positive
Review: if you love reading and\/or poetry , then by all means check it out . Sentiment: positive
Review: -lrb- lee -rrb- treats his audience the same way that jim brown treats his women -- as dumb , credulous , unassuming , subordinate subjects . Sentiment: negative
Review: a fairly by-the-books blend of action and romance with sprinklings of intentional and unintentional comedy .
Prediction: positive
\end{lstlisting}

\paragraph{MMR retrieved prompt.}\mbox{}\\[-0.8em]
\begin{lstlisting}[style=prompt]
Review: an odd , haphazard , and inconsequential romantic comedy . Sentiment: negative
Review: the ya-ya 's have many secrets and one is - the books are better . Sentiment: negative
Review: the ensemble cast turns in a collectively stellar performance , and the writing is tight and truthful , full of funny situations and honest observations . Sentiment: positive
Review: behan 's memoir is great material for a film -- rowdy , brawny and lyrical in the best irish sense -- but sheridan has settled for a lugubrious romance . Sentiment: negative
Review: the most repugnant adaptation of a classic text since roland joffe and demi moore 's the scarlet letter . Sentiment: negative
Review: a fairly by-the-books blend of action and romance with sprinklings of intentional and unintentional comedy .
Prediction: negative
\end{lstlisting}

\paragraph{Top-$k$ retrieved prompt.}\mbox{}\\[-0.8em]
\begin{lstlisting}[style=prompt]
Review: an odd , haphazard , and inconsequential romantic comedy . Sentiment: negative
Review: the story that emerges has elements of romance , tragedy and even silent-movie comedy . Sentiment: positive
Review: a witty , low-key romantic comedy . Sentiment: positive
Review: a smart , sassy and exceptionally charming romantic comedy . Sentiment: positive
Review: what sets this romantic comedy apart from most hollywood romantic comedies is its low-key way of tackling what seems like done-to-death material . Sentiment: positive
Review: a fairly by-the-books blend of action and romance with sprinklings of intentional and unintentional comedy .
Prediction: negative
\end{lstlisting}

\subsection{MathQA}
\paragraph{Gold label.} c.

\paragraph{RES retrieved prompt.}\mbox{}\\[-0.8em]
\begin{lstlisting}[style=prompt]
Question: a man bought 40 shares of rs . 60 at 5 discount , the rate of dividend being 1212 % the rate of interest obtained is
Options:
a ) 13.64 % , b ) 15.5 % , c ) 14 % , d ) 14.25 % , e ) 14.95 %
Answer: a
Explanation: explanation : face value of a share = rs . 60 he bought each share at rs . 60 - rs . 5 = rs . 55 number of shares = 40 dividend = 12 1 / 2 % = 25 / 2 % dividend per share = 60 × 25 / 2 × 100 = rs . 7.5 total dividend = ( 40 × 7.5 ) ie , he got a dividend of ( 40 × 7.5 ) for an investment of rs . ( 40 × 55 ) interest obtained = 40 × 7.5 × 100 / 40 × 55 = 13.64 % answer : option a
Question: find the number of zeroes at the end of 50 !
Options:
a ) 10 , b ) 11 , c ) 12 , d ) 13 , e ) 14
Answer: c
Explanation: "calculate division by 5 and 25 50 / 5 = 10 50 / 25 = 2 total = 12 answer : c"
Question: if $ 5,000 is invested in an account at a simple annual rate of r percent , the interest is $ 250 . when $ 19,000 is invested at the same interest rate , what is the interest from the investment ?
Options:
a ) $ 700 , b ) $ 750 , c ) $ 800 , d ) $ 950 , e ) $ 900
Answer: d
Explanation: "- > 250 / 5,000 = 5 % and 19,000 * 5 % = 950 . thus , d is the answer ."
Question: solve the equation for x : 19 ( x + y ) + 17 = 19 ( - x + y ) - 21 ?
Options:
a ) - 1 , b ) 77 , c ) 8 , d ) 3 , e ) 2
Answer: a
Explanation: "19 x + 19 y + 17 = - 19 x + 19 y - 21 38 x = - 38 = > x = - 1 answer : a"
Question: ravi and sunil are partners in a business . ravi invests rs . 20,000 for 8 months and sunil invested rs . 16000 for 10 months then after one year ratio of their profits will be
Options:
a ) 1 : 1 , b ) 2 : 3 , c ) 3 : 2 , d ) 3 : 1 , e ) 3 : 4
Answer: a
Explanation: "= ( 15000 * 8 ) : ( 8000 * 10 ) = 160000 : 160000 = 1 : 1 answer : a"
Question: a man bought 20 shares of rs . 50 at 5 discount , the rate of dividend being 13 . the rate of interest obtained is :
Options:
a ) 27 % , b ) 87 % , c ) 15 % , d ) 66 % , e ) 88 %
Prediction: c
\end{lstlisting}

\paragraph{MMR retrieved prompt.}\mbox{}\\[-0.8em]
\begin{lstlisting}[style=prompt]
Question: a man bought 40 shares of rs . 60 at 5 discount , the rate of dividend being 1212 % the rate of interest obtained is
Options:
a ) 13.64 % , b ) 15.5 % , c ) 14 % , d ) 14.25 % , e ) 14.95 %
Answer: a
Explanation: explanation : face value of a share = rs . 60 he bought each share at rs . 60 - rs . 5 = rs . 55 number of shares = 40 dividend = 12 1 / 2 % = 25 / 2 % dividend per share = 60 × 25 / 2 × 100 = rs . 7.5 total dividend = ( 40 × 7.5 ) ie , he got a dividend of ( 40 × 7.5 ) for an investment of rs . ( 40 × 55 ) interest obtained = 40 × 7.5 × 100 / 40 × 55 = 13.64 % answer : option a
Question: find the number of zeroes at the end of 50 !
Options:
a ) 10 , b ) 11 , c ) 12 , d ) 13 , e ) 14
Answer: c
Explanation: "calculate division by 5 and 25 50 / 5 = 10 50 / 25 = 2 total = 12 answer : c"
Question: what is the remainder when the number r = 14 ^ 2 * 15 ^ 8 is divided by 5 ?
Options:
a ) 0 , b ) 1 , c ) 2 , d ) 4 , e ) 5
Answer: a
Explanation: "14 ^ 2 has units digit 6 15 ^ 8 has units digit 5 thus r = 14 ^ 2 * 15 ^ 8 has units digit 0 and will be divisible by 5 . the remainder will be zero answer : ( a )"
Question: if the numerator of a fraction is increased by 50 % and denominator by 20 % , the fraction become 9 / 15 . what was the original fraction ?
Options:
a ) 12 / 25 , b ) 13 / 25 , c ) 3 / 5 , d ) 6 / 7 , e ) 7 / 8
Answer: a
Explanation: "explanation : let the fraction be x / y we have 150 x / 120 y = 9 / 15 = x / y = 12 / 25 answer : option a"
Question: the difference b / w the compound interest payble half yearly and the simple interest on a certain sum lent out at 10 % p . a for 1 year is rs 25 . what is the sum ?
Options:
a ) rs . 15000 , b ) rs . 12000 , c ) rs . 10000 , d ) rs . 20000 , e ) none of these
Answer: c
Explanation: in simple words s . i = ptr / 100 = ( p * 1 * 10 ) / 100 p / 10 here amount a = p + s . i = p + p / 10 = 11 p / 10 . . . . now for compound interest when calculated half yearly amount = p [ 1 + r / 200 ] ^ 2 t = p [ 1 + 10 / 200 ] ^ 2 * 1 = 441 p / 400 here its given in the question dat diff btwn simple interest and compound interest half yearly . . . is 25 = > 441 p / 400 - 11 p / 10 = 25 = > p / 400 = 25 = > p = 400 * 25 = 10000 so sum = 10000 answer : c
Question: a man bought 20 shares of rs . 50 at 5 discount , the rate of dividend being 13 . the rate of interest obtained is :
Options:
a ) 27 % , b ) 87 % , c ) 15 % , d ) 66 % , e ) 88 %
Prediction: a
\end{lstlisting}

\subsection{AQUA}
\paragraph{Gold label.} A.

\paragraph{RES retrieved prompt.}\mbox{}\\[-0.8em]
\begin{lstlisting}[style=prompt]
Question: The probability that event A occurs is 0.4, and the probability that events A and B both occur is 0.25. If the probability that either event A or event B occurs is 0.8, what is the probability that event B will occur?
Options:
A)0.05
B)0.15
C)0.45
D)0.5
E)0.65
Answer: E
Explanation: P(A or B) = P (A) + P(B) - p(a n b)
0.8= 0.4 + P(B) - 0.25
P(B) = 0.65
Ans : E
Question: Find the fraction which has the same ratio to 2/6 that 3/4 has to 1/2
Options:
A)1/2
B)2/5
C)6/8
D)9/4
E)7/5
Answer: A
Explanation: P : 2/6 = 3/4 : 1/2
As the product of the means is equal to the product of the extremes.
P*1/2 = 2/6 * 3/4
P*1/2 = 6/24
P = 1/2 => P = 1/2
ANSWER:A
Question: If f(a)=a^2, what is the value of (f(a+b)+f(a))/b ?
Options:
A)a
B)b
C)2a
D)2a + b
E)2a+b+2a^2
Answer: E
Explanation: start with f(a+b) before start calculations: f(a+b)=(a+b)^2= a^2+2ab+b^2
a^2+2ab+b^2 + a^2/b = (simplify a^2)
2ab+b^2/b+2a^2 = (cancel b's)
= 2a+b+2a^2
Answer : E
Question: W and X started a business investing Rs. 30,000 and Rs. 6,000 respectively. Out of a total profit of Rs. 18,000, W’s share is :
Options:
A)14000
B)15000
C)16000
D)17000
E)18000
Answer: B
Explanation: Ratio of their shares = 30000 : 6000 = 5 : 1.
W’s share = Rs. 18000 * 5/6
= Rs. 15000
ANSWER:B
Question: loga (ab) = x, then log b (ab) is :
Options:
A)1/x
B)x/ (x+1)
C)x/(1-x)
D)x/(x-1)
E)x/(x-2)
Answer: D
Explanation: loga (ab) = x => log b/ log a = x => (log a + log b)/ log a = x
1+ (log b/ log a) = x => log b/ log a = x-1
log a/ log b = 1/ (x-1) => 1+ (log a/ log b) = 1 + 1/ (x-1)
(log b/ log b) + (log a/ log b) = x/ (x-1) => (log b + log a)/ log b = x/ (x-1)
=>log (ab)/ log b = x/(x-1) => logb (ab) = x/(x-1)
ANSWER:D
Question: Let A and B be independent events with P(A)=0.2 and P(B)=0.8. Find P(A/B)?
Options:
A)0.2
B)0.4
C)0.6
D)1.2
E)1.5
Prediction: B
\end{lstlisting}

\paragraph{MMR retrieved prompt.}\mbox{}\\[-0.8em]

\begin{lstlisting}[style=prompt]
Question: The probability that event A occurs is 0.4, and the probability that events A and B both occur is 0.25. If the probability that either event A or event B occurs is 0.8, what is the probability that event B will occur?
Options:
A)0.05
B)0.15
C)0.45
D)0.5
E)0.65
Answer: E
Explanation: P(A or B) = P (A) + P(B) - p(a n b)
0.8= 0.4 + P(B) - 0.25
P(B) = 0.65
Ans : E
Question: What is the reciprocal of (AB)/(A+B)^2?
Options:
A)(AB)/(A^2+2AB+B^2)
B)A/B+B/A+AB
C)B/A+A/B+2
D)(A+B)^2/AB
E)A/B+2AB
Answer: C
Explanation: Answer:
Reciprocal = (A+B)^2/AB
= (A^2+B^2+2AB)/AB = B/A + A/B +2
ANSWER:C
Question: If b does not equal zero, and ab = b/2, what is the value of a?
Options:
A)A)1/8
B)B)1/4
C)C)1/3
D)D)1/2
E)of the above
Answer: D
Explanation: Explanation:
To solve for a, divide both sides of the equation by b: ab = b/2
(ab)/b = (b/2)/b
a = (b/2)*1/b
a = 1/2
Answer: (D).
Question: If P(A) = 3/5 and P (B) = 2/5, find P (A n B) if A and B are independent events.
Options:
A)6/25
B)3/25
C)8/25
D)2/13
E)3/17
Answer: A
Explanation: P (A n B) = P(A) . P(B)
P (A n B) = 3/5 . 2/5
P (A n B) = 6/25.
A
Question: The probability of pulling a black ball out of a glass jar is 1/X. The probability of pulling a black ball out of a glass jar and breaking the jar is 1/Y. What is the probability of breaking the jar?
Options:
A)1/(XY)
B)X/Y
C)Y/X
D)1/(X+Y)
E)1/(X-Y)
Answer: B
Explanation: Probability of A and B:
When two events are independent, the probability of both occurring is the product of the probabilities of the individual events: P(A and B)=P(A)*P(B)P(A and B)=P(A)*P(B)
P(break & black) = P(break) * P(black) --> 1/y=P(break)*1/x --> P(break)=x/y.
Answer: B.
Question: Let A and B be independent events with P(A)=0.2 and P(B)=0.8. Find P(A/B)?
Options:
A)0.2
B)0.4
C)0.6
D)1.2
E)1.5
Prediction: A
\end{lstlisting}
\end{document}